\documentclass[10pt,twocolumn,letterpaper]{article}

\usepackage[pagenumbers]{cvpr} 

\usepackage[dvipsnames]{xcolor}

\definecolor{cvprblue}{rgb}{0.21,0.49,0.74}
\usepackage[pagebackref,breaklinks,colorlinks,citecolor=cvprblue]{hyperref}

\usepackage{times}
\usepackage{epsfig}
\usepackage{graphicx}
\usepackage{amsmath}
\usepackage{amssymb}

\usepackage{booktabs}
\usepackage{commath}
\usepackage{mathtools}
\usepackage{enumerate}
\usepackage{enumitem}
\usepackage{xcolor}
\usepackage{soul}

\usepackage{caption}
\usepackage{subcaption}
\usepackage[font=small,skip=3pt]{caption}
\usepackage{physics}
\usepackage{algorithm}
\usepackage{algcompatible}
\usepackage[dvipsnames]{xcolor}
\usepackage[accsupp]{axessibility}

\usepackage[utf8]{inputenc}
\algnewcommand\INPUT{\item[\textbf{Input:}]}%
\algnewcommand\INIT{\item[\textbf{Init:}]}%
\algnewcommand\OUTPUT{\item[\textbf{Output:}]}%

\def\paperID{12421} 
\def\confName{CVPR}
\def\confYear{2024}

\title{A Dual-Augmentor Framework for Domain Generalization in 3D Human Pose Estimation}

\author{Qucheng Peng, Ce Zheng, Chen Chen\\
Center for Research in Computer Vision,
University of Central Florida\\
{\tt\small \{qucheng.peng,ce.zheng\}@ucf.edu, chen.chen@crcv.ucf.edu}\\
}

\begin{document}
\maketitle
\begin{abstract}
3D human pose data collected in controlled laboratory settings present challenges for pose estimators that generalize across diverse scenarios. To address this, domain generalization is employed. Current methodologies in domain generalization for 3D human pose estimation typically utilize adversarial training to generate synthetic poses for training. Nonetheless, these approaches exhibit several limitations. First, the lack of prior information about the target domain complicates the application of suitable augmentation through a single pose augmentor, affecting generalization on target domains. Moreover, adversarial training's discriminator tends to enforce similarity between source and synthesized poses, impeding the exploration of out-of-source distributions. Furthermore, the pose estimator's optimization is not exposed to domain shifts, limiting its overall generalization ability.

To address these limitations, we propose a novel framework featuring two pose augmentors: the weak and the strong augmentors. Our framework employs differential strategies for generation and discrimination processes, facilitating the preservation of knowledge related to source poses and the exploration of out-of-source distributions without prior information about target poses. Besides, we leverage meta-optimization to simulate domain shifts in the optimization process of the pose estimator, thereby improving its generalization ability. Our proposed approach significantly outperforms existing methods, as demonstrated through comprehensive experiments on various benchmark datasets. Our code will be released at \url{https://github.com/davidpengucf/DAF-DG}.
\end{abstract}    
\section{Introduction}
\label{sec:intro}

\begin{figure}[!ht]
  \centering
   \includegraphics[width=1.0\linewidth]{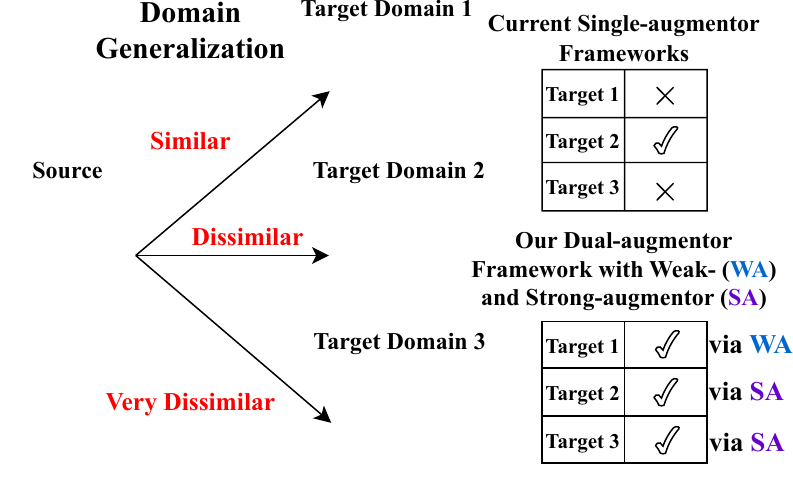}\vspace{-0mm}
   \caption{Comparisons between existing single-augmentor frameworks and our proposed dual-augmentor framework on a toy example. Current single-augmentor methods excel at simulating Target Domain 2 but exhibit limitations in simulating Target Domain 1, closely resembling the source, and Target Domain 3, deviating significantly from the source. In our framework, the weak augmentor excels in simulating Target Domain 1, while the strong augmentor effectively imitates both Target Domain 2 and 3.
   }\vspace{-10pt}
   \label{fig:state}
\end{figure}

3D human pose estimation (HPE) is the process of predicting the 3D coordinates of human joints from images or videos. It serves as the foundation for various applications including person re-identification \cite{su2017pose}, action recognition \cite{hua2023part,lu2023hard,yan2023mae}, human mesh recovery \cite{zheng2023feater,zheng2023potter}, virtual reality \cite{guzov2021human,yi2023mime}. 
However, the annotated 3D data are often collected in controlled laboratory environments for convenience, featuring indoor settings and limited actions performed by few individuals. As a result, pose estimators trained on these labeled datasets face challenges in generalizing to varied in-the-wild scenarios. Hence, the notion of domain generalization (DG) is pivotal in incorporating knowledge from labeled (source) data into a pose estimator that could generalize well on unseen (target) data. Unlike domain adaptation (DA) which involves the training with target data, \ul{DG relies solely on the source data as a reference, without any prior information about the target data.}

Existing DG approaches for 3D HPE \cite{zhang2023poseaug,huang2022dh,guan2023posegu} conduct substantial augmentations on the source poses to obtain synthesized poses via adversarial training, and incorporate the synthesized poses as complementary to the source poses for HPE model training. However, these methods have several limitations. \emph{First}, in the context of DG for 3D HPE, there is a complete lack of information about poses in target domains. If target poses closely resemble the source (as Target Domain 1 in Fig. \ref{fig:state}), poses generated by extensive augmentation notably differ from the target, thereby impeding generalization. Conversely, when target poses significantly deviate from the source distributions (as Target Domain 3 in Fig. \ref{fig:state}), poses generated via insufficient augmentation may not sufficiently explore out-of-source knowledge to simulate the target. \ul{Existing methods only use a single augmentor, making it challenging to simultaneously achieve both objectives.} \emph{Second}, the adversarial training between synthesized and source poses constrains the diversity of generation. Existing methods typically employ the generative adversarial network (GAN) \cite{goodfellow2014generative} structure, which includes one pose generator responsible for pose generation and one discriminator to assist the pose generator by providing feedback. Specifically, the discriminator enforces similarity between synthesized and source poses, aiming to ensure that the generated poses closely resemble the source poses, which harms the exploration of out-of-source knowledge.  

To address these limitations, we propose a novel framework featuring two pose augmentors: the weak augmentor and the strong augmentor. The weak augmentor is designed to simulate target poses closely resembling source poses, while the strong augmentor generates target poses that exhibit significant deviations from source distributions. To delineate their characteristics, differential strategies are employed for generation and discrimination processes, as detailed in Sec. \ref{sec:aug}. Notably, our framework alleviates the constraints on strong-augmented poses by traditional adversarial training methods. Instead of enforcing similarity between the source and all the augmented poses, we utilize weak-augmented poses as an intermediary, enabling discrimination between strong- and weak-augmented poses and facilitating discrimination between source and weak-augmented poses. 
To optimize the utilization of synthesized poses, we introduce meta-optimization among source, weak-augmented, and strong-augmented poses, as elaborated in Sec. \ref{sec:meta}. Our training process exposes the pose estimator to domain shifts during the optimization processes, thereby enhancing its adaptability to handle domain shifts during the inference stage.
Our contributions can be summarized in three main aspects:

\setlist{nolistsep}
\begin{itemize}[noitemsep,leftmargin=*] 
    \item We propose a novel framework featuring both the weak and strong pose augmentors, which effectively preserves knowledge related to source poses while simultaneously exploring out-of-source distributions through differential strategies for the generation and discrimination processes of the two augmentors.
    \item We introduce meta-optimization to enhance the utilization of synthesized poses. By simulating domain shifts among source, weak-augmented, and strong-augmented poses during the optimization processes, the pose estimator's generalization ability is further improved. 
    \item We conduct comprehensive experiments on several benchmark datasets, and the results demonstrate that our approach significantly outperforms state-of-the-art methods by a considerable margin. 
\end{itemize}

\section{Related Work}
\label{sec:related}
\textbf{3D Human Pose Estimation.} The widely adopted two-stage technique in 3D HPE, as demonstrated in \cite{zhao2019semgcn,pavllo20193d,zheng20213d,zhang2022mixste,zhao2023poseformerv2,zhu2023motionbert}, initially employs 2D human pose estimators \cite{li2020cascaded,wang2020deep,peng2023source,pinyoanuntapong2023gaitsada} for 2D pose predictions, followed by lifting these predictions to 3D poses. Among these approaches, \cite{zhao2019semgcn} integrates graph-structured semantic information to enhance the estimation process, while \cite{pavllo20193d} utilizes dilated temporal convolutional layers for temporal information encoding, and \cite{zheng20213d} presents a purely transformer-based 3D approach. Moreover, \cite{zhang2022mixste} effectively models inter-frame correspondences with a mixed sequence-to-sequence encoder, and recent works such as \cite{zhao2023poseformerv2} explore the frequency domain to improve inference efficiency, and \cite{zhu2023motionbert} employs unified representation learning for 3D human poses.


\noindent\textbf{Domain Generalization.} Current DG methods aim to learn domain-invariant representations and are categorized into three types: domain alignment \cite{zhao2020domain,matsuura2020domain}, meta-learning \cite{li2018learning,qiao2020learning}, and augmentation strategies \cite{zhao2020maximum,zhao2022style,peng2023rain,peng2022multi}. For domain alignment, \cite{zhao2020domain} enhances the conditional invariance of learned features by incorporating an entropy regularization term, leading to improved classifier generalization. \cite{matsuura2020domain} iteratively segregates samples into latent domains through clustering. Concerning meta-learning, \cite{li2018learning} proposes a model-agnostic training procedure that simulates domain shift during training, whereas \cite{qiao2020learning} applies meta-learning to single-domain generalization. Regarding augmentation strategies, \cite{zhao2020maximum} introduces a novel regularization term for adversarial data augmentation derived from the information bottleneck principle, while \cite{zhao2022style} presents a unique style hallucination module to generate style-diversified samples crucial for generalization.

Differing from the current DG approaches for 3D HPE that focus solely on augmentations, we also incorporate meta-learning-based approaches to enhance generalization.

\noindent\textbf{Cross-domain Learning for 3D Human Pose Estimation.} Cross-domain learning for 3D HPE is categorized into two types: \emph{domain generalization} \cite{zhang2023poseaug,huang2022dh,guan2023posegu,li2023cee} and \emph{domain adaptation} \cite{gholami2022adaptpose,chai2023poseda,liu2023posynda}. \ul{In domain generalization, training processes exclusively utilizes source data, and the resulting model is directly applied to infer target data.} 
\cite{zhang2023poseaug} adjusts various geometry factors of human poses through differentiable operations. \cite{huang2022dh} applies DH Forward Kinematics \cite{baili2004classification} to drive 3D pose augmentation and obtain diverse poses. \cite{guan2023posegu} incorporates Counterfactual Risk Minimization \cite{swaminathan2015counterfactual} to achieve unbiased learning. \cite{li2023cee} addresses with network designs like the interpolation sub-net and body-parts grouping net. \ul{In domain adaptation, labeled source data and unlabeled target data are used simultaneously during the training process.} \cite{gholami2022adaptpose} employs generative adversarial network \cite{goodfellow2014generative} to discriminate between source and target during training. \cite{chai2023poseda} utilizes global position alignment and local pose augmentation to transfer from source to target. \cite{liu2023posynda} employs a multi-hypotheses network along with target-specific source augmentation for the problem.

In this paper, \textbf{we focus on domain generalization for 3D HPE}. In addition to synthesizing novel 3D poses for better generalization, we also simulate domain shifts by using both source and synthesized poses during optimizations.

\section{Methodology}
\label{sec:method}

\begin{figure*}[!ht]
  \centering
   \includegraphics[width=1.0\linewidth]{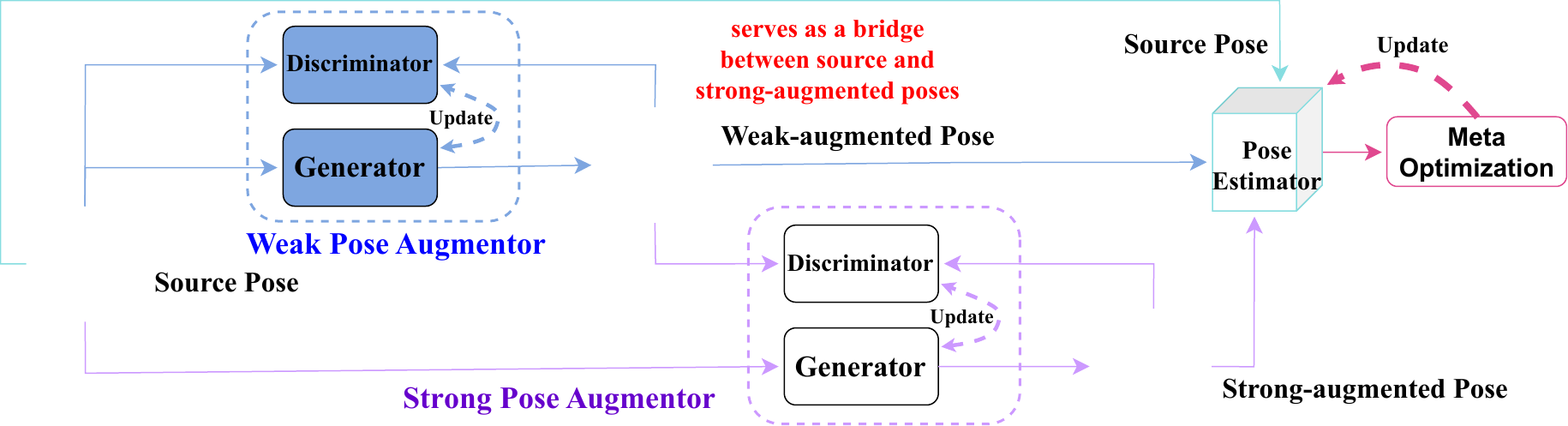}\vspace{-0mm}
   \caption{Overall framework of our dual-augmentor method. Initially, the original pose undergoes processing through two pose augmentors, resulting in weak- and strong-augmented poses (See Sec. \ref{sec:aug}). The weak augmentor simulates target domains similar to the source domain, while the strong augmentor emulates target domains that deviate significantly from the source distributions. Subsequently, the original pose and the two augmented poses are input to the pose estimator for further meta-optimization (See Sec. \ref{sec:meta}).
   }\vspace{-10pt}
   \label{fig:framework}
\end{figure*}

\subsection{Preliminaries}
\noindent\textbf{2D-to-3D lifting Paradigm of 3D HPE.} Current 2D-to-3D lifting paradigm of 3D HPE \cite{pavllo20193d,zheng20213d,zhang2022mixste,zhao2019semgcn} assumes that $x^{s}_{i} \in \mathbb{R}^{J \times 2}$ represents the 2D coordinates of $J$ keypoints of a sample in the source domain (2D poses as input), and $y^{s}_{i} \in \mathbb{R}^{J \times 3}$ represents the corresponding 3D positions in the camera coordinate system (3D poses as output), we denote the source domain with $N$ samples as $S = \{(x_{i}^{s}, y_{i}^{s})\}_{i=1}^N$, encompassing $N$ 2D-3D pairs. Moreover, we define the pose estimator  as $\mathcal{P}: x^{s}_{i} \mapsto \hat{y}_{i}^{s}$, where $\hat{y}_{i}^{s}$ represents the predicted corresponding 3D pose positions. For a fully supervised human pose estimation problem, we aim to achieve an ideal $\mathcal{P}$ by solving the following optimization objective:
\begin{equation}
\min_{\mathcal{P}} \mathbb{E}_{(x^{s}_{i},y^{s}_{i}) \in \mathcal{S}} \mathcal{L}_{MSE}(\mathcal{P}(x_{i}^{s}), y^{s}_{i}),
\label{eq:sup}
\end{equation} where $\mathcal{L}_{MSE}$ represents the Mean Squared Error (MSE) loss. However, the objective \cite{pavllo20193d,zheng20213d} is designed to achieve optimal performance on source poses, rendering it inadequate for addressing the DG problem, as it does not account for the domain gap between source and target domains.

\noindent\textbf{DG for 3D HPE.} Within the paradigm of 2D-to-3D lifting HPE, our primary goal is to derive an estimator $\mathcal{P}$ that demonstrates commendable 3D HPE performance specifically within the target domain $T$. Under this scenario, the target domain $T = \{(x_{j}^{t}, y_{j}^{t})\}_{j=1}^M$ with $M$ samples can only be used for inference and is not involved in the training process. However, when utilizing solely the original source domain, the pose estimator cannot learn out-of-source distributions, which is essential for achieving good performance on the target domain. Existing methods \cite{zhang2023poseaug,huang2022dh,li2023cee,guan2023posegu} tend to conduct augmentation to the original source poses to enhance data diversity, thereby improving the estimator's generalization ability. The augmentor is denoted as $\mathcal{A}: y^{s}_{i} \mapsto {y}_{i}^{a}$, while the projection from 3D to 2D via camera parameters (completely known) is defined as $\mathcal{R}: y^{a}_{i} \mapsto {x}_{i}^{a}$. Consequently, the min-max optimization objective for domain generalization can be defined as follows:
\begin{equation}
\begin{multlined}
\min_{\mathcal{P}} \max_{\mathcal{A}} \mathbb{E}_{(x^{s}_{i},y^{s}_{i}) \in \mathcal{S}} [\mathcal{L}_{MSE}(\mathcal{P}(x_{i}^{s}), y^{s}_{i}) \\ +  \mathcal{L}_{MSE}(\mathcal{P}(\mathcal{R}(\mathcal{A}(y^{s}_{i}))),\mathcal{A}(y^{s}_{i}))].
\label{eq:dg}
\end{multlined}
\end{equation}
The objective is a \textbf{min-max game} between the pose augmentor $\mathcal{A}$ and the pose estimator $\mathcal{P}$, encouraging the estimator $\mathcal{P}$ to learn out-of-source distributions, while conducting augmentations to a significant extent is beneficial to generate more diverse samples, and that is why the loss is minimized with respect to the augmentor $\mathcal{P}$ and maximized with respect to the augmentor $\mathcal{A}$ in the optimization.

\subsection{Overview of the Proposed Method}

Existing methods \cite{zhang2023poseaug,huang2022dh,guan2023posegu} often apply intricate augmentations to the original poses in the source domain, relying on the discrimination between augmented poses and source poses simultaneously. However, this kind of approach raises two concerns. \emph{First}, as this is a DG problem for 3D HPE, any information about the target domain is entirely unknown. If the target domain bears a striking resemblance to the source domain (like Target Domain 1 in Fig. \ref{fig:state}), poses generated by extensive augmentation might hinder the pose estimator's inference on it. Conversely, in cases where the target domain significantly diverges from the source distributions (like Target Domain 3 in Fig. \ref{fig:state}), poses generated by insufficient augmentation may fail to adequately explore out-of-source knowledge for the pose estimator. \emph{Second}, when target domain is distant from the source and needs significant augmentations, the adversarial training between source and synthesized poses limits the diversity of generated poses. Specifically, the discriminator enforces similarity between source and synthesized poses, 
thereby causing the synthesized poses to remain similar to the source poses. 

To tackle these concerns, we propose a novel dual-augmentor framework depicted in Fig. \ref{fig:framework}. \textbf{This framework involves two augmentors that generate weak- and strong-augmented poses}, enabling the handling of diverse unknown target domains. Additionally, the weak-augmentation module serves as a bridge between strong-augmented and source poses. Specifically, \textcolor{blue}{the discrimination between source poses and weak-augmented poses is utilized to update the weak augmentor}, while \textcolor{Plum}{the discrimination between weak- and strong-augmented poses is employed to optimize the strong augmentor}. This approach liberates the strong augmentor from heavy reliance on the source domain and enables the exploration of more out-of-source knowledge. Further details regarding the pose augmentation process can be found in Sec. \ref{sec:aug}.

Having elucidated the methodology for synthesizing poses, the subsequent discourse pivots towards the utilization of these synthesized poses. Previous works \cite{zhang2023poseaug,huang2022dh,guan2023posegu} overlook the interactions between source poses and augmented poses, dealing with the optimizations of pose estimator on them separately. 
In contrast, we propose a model-agnostic meta optimization approach that enhances the interactions between source poses and the two types of augmented poses to simulate domain shifts in the optimization and leverage domain-invariant knowledge while maintaining the original 2D-to-3D lifting backbone's structure unchanged. Further details concerning the meta optimization process can be found in Sec. \ref{sec:meta}.

\subsection{Pose Augmentation}
\label{sec:aug}

\begin{figure}[ht]
\vspace{-10pt}
  \centering
   \includegraphics[width=1.0\linewidth]{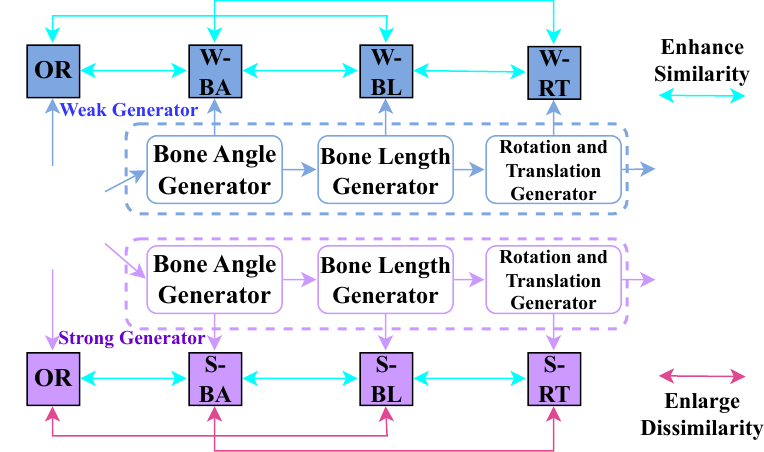}
   \caption{The differentiation of the weak and strong generators. Within each pipeline, denoted as "W-" for weak ones and "S-" for strong ones, there exist four pose states: original (OR), after bone angle operation (BA), after bone length operation (BL) and after rotation and translation operation (RT). For proximate states, similarities are enhanced for both generators. When there is a one-state gap between states, the weak generator continues to enhance similarities, whereas the strong generator enlarges dissimilarities.
   }\vspace{-5pt}
   \label{fig:augmentation}
\end{figure}

The pose augmentation architecture comprises two pose augmentors: the weak augmentor and the strong augmentor. Each augmentor comprises two components: the generator, responsible for producing diverse synthesized poses to facilitate the training of the pose estimator, and the discriminator, which collaborates with the generator to regulate the quality of the generated poses. Our objective is to apply \emph{differential strategies} to the generation (named \textbf{differential generation}) and discrimination (named \textbf{differential discrimination}) of the two augmentors, enabling them to generate weak- and strong-augmented poses. 

\noindent\textbf{Differential Generation:} The generation process, as illustrated in Figure \ref{fig:augmentation}, consists of three modules in sequence for each generator pipeline: the Bone Angle Generator, Bone Length Generator, and Rotation and Translation Generator, resulting in four statuses in the pipeline: original (OR), after the bone angle operation (BA), after the bone length operation (BL), and after the rotation and translation operation (RT). Existing approaches such as \cite{zhang2023poseaug,li2023cee} typically treat the entire generation pipeline in an end-to-end manner and only utilize the $(OR, RT)$ pair. In contrast, our method deals with the generation in a more fine-grained fashion. We group statuses into pairs based on their relations: proximate pairs as $PP = \{(OR, BA), (BA, BL), (BL, RT)\}$, and one-state gap pairs as $OG = \{(OR, BL), (BA, RT)\}$.

To begin, we define the measurement of similarity within a pair. Solely relying on conventional absolute position losses, such as the MSE loss, is not adequate in this context for two reasons. First, the three modules within the generator all perform operations on the level of bone vector, not on the joint positions. If one joint undergoes significant position changes after an operation, other joints connected to it will also experience considerable movement, even if the bone vector between them remains stable. In such cases, position-based measurements cannot fully reflect the extent of augmentation based on the bone vector. Second, human poses possess kinematic attributes, and a position-based measurement overlooks the graphical information. Therefore, we introduce the Laplacian weighted similarity measurement. For the human model, it is straightforward to obtain degree matrix $D$ and adjacency matrix $A$, and the normalized Laplacian matrix can be represented as:
\begin{equation}
    W_{NL} = I - D^{-\frac{1}{2}}AD^{-\frac{1}{2}},
\label{eq:laplacian}
\end{equation} where $I$ is the identity matrix, and $W_{NL}$ is the normalized Laplacian matrix encoding graphical information. Given a pair of statuses $(st_{1}, st_{2})$ (from either PP or OG), the similarity measurement is defined as:
\begin{small}
\begin{equation}
    \mathcal{L}_{sim}(st_1, st_2) = \underbrace{\lVert st_1 - st_2\rVert}_\text{MSE Similarity} + \underbrace{\lVert W_{NL}st_1 - W_{NL}st_2\rVert}_\text{Laplacian Weighted Similarity}.
\label{eq:sim}
\end{equation}
\end{small}

To differentiate between the two generators, we apply distinctive strategies. For the weak generator, we enhance similarities for its PP and OG sets to maintain a \emph{slight} level of augmentation in the synthesized poses, as indicated by:
\begin{equation}
\begin{multlined}
    \mathcal{L}_{wg} = \mathop{\mathbb{E}}_{(st_1,st_2) \in PP} \mathcal{L}_{sim}(st_1, st_2) \\+  \alpha_1 \mathop{\mathbb{E}}_{(st_1,st_2) \in OG} \mathcal{L}_{sim}(st_1, st_2).
\label{eq:weak_gen}
\end{multlined}
\end{equation} For the strong generator, we enhance similarities within its PP set to ensure the reasonableness of the synthesized output, while enlarging dissimilarities within its OG sets to maintain a \emph{significant} level of augmentation, expressed as:
\begin{equation}
\begin{multlined}
    \mathcal{L}_{sg} = \mathop{\mathbb{E}}_{(st_1,st_2) \in PP} \mathcal{L}_{sim}(st_1, st_2) \\-  \alpha_2 \mathop{\mathbb{E}}_{(st_1,st_2) \in OG} \mathcal{L}_{sim}(st_1, st_2).
\label{eq:strong_gen}
\end{multlined}
\end{equation} $\alpha_1$ and $\alpha_2$ are trade-off parameters. 

\noindent\textbf{Differential Discrimination:} The discrimination process comprises two min-max games \cite{goodfellow2014generative,gulrajani2017improved}: one between the source pose and the weak-augmented poses, and the other between the weak-augmented pose and the strong-augmented pose. We adopt the WGAN-GP \cite{gulrajani2017improved} structure here. The discrimination losses regarding the source poses $y^{or}$, weak-augmented poses $y^{wa}$, and strong-augmented poses $y^{sa}$ are defined as follows:
\begin{equation}
\begin{multlined}
    \mathcal{L}_{wd} = \mathbb{E}[D_{wa}(y^{or})] - \mathbb{E}[D_{wa}(y^{wa})] \\ + \beta_1 \mathbb{E}(1 - \lVert \nabla_{\hat{y}^{wa}}D_{wa}(\hat{y}^{wa})  \rVert),
\label{eq:weak_dis}
\end{multlined}
\end{equation}
\begin{equation}
\begin{multlined}
    \mathcal{L}_{sd} = \mathbb{E}[D_{sa}(y^{wa})] - \mathbb{E}[D_{sa}(y^{sa})] \\ + \beta_2 \mathbb{E}(1 - \lVert \nabla_{\hat{y}^{sa}}D_{sa}(\hat{y}^{sa})  \rVert).
\label{eq:strong_dis}
\end{multlined}
\end{equation}
Here, $\mathcal{L}_{wd}$ is the discrimination loss between $y^{or}$ and $y^{wa}$, used to update the weak augmentor, and $D_{wa}$ is the weak discriminator. $\mathcal{L}_{sd}$ is the discrimination loss between $y^{wa}$ and $y^{sa}$, used to update the strong augmentor, and $D_{sa}$ is the strong discriminator. $\hat{y}^{wa}$ and $\hat{y}^{sa}$ are built via interpolation, such that $\hat{y}^{wa} = \epsilon y^{or} + (1-\epsilon)y^{wa}$ and $\hat{y}^{sa} = \epsilon y^{wa} + (1-\epsilon)y^{sa}$, where $\epsilon$ is randomly drawn from $U[0,1]$. $\beta_1$ and $\beta_2$ are trade-off parameters.

By implementing this discrimination process in two min-max games, the weak augmentor is capable of retaining more source information and alleviating adverse effects stemming from irrationally synthesized poses. Simultaneously, the strong augmentor can overcome a strong dependency on the source distributions, and explore out-of-source distributions more effectively. With diverse synthesized poses to simulate potential target poses, it is beneficial for further domain generalization in pose estimation.

\subsection{Meta Optimization}
\label{sec:meta}

\vspace{-10pt}
\begin{algorithm}[!ht]
  \caption{Meta Optimization Pseudo Code}
  \begin{algorithmic}[1]
    \INPUT Original source 2D-3D pose pairs $(x^{or}, y^{or})$; Weak Generator $G_{wa}$; Strong Generator $G_{sa}$
    \INIT Pose estimator $\mathcal{P}_{t}$, Learning rates $lr_1$ and $lr_2$, inner loop iteration $k$, Hyperparameter $\gamma$  
    \OUTPUT Updated pose estimator $\mathcal{P}_{t+2}$ after two-step meta optimization 
    \STATE Generate weak-augmented 3D poses \begin{small}$y^{wa} = G_{wa}(y^{or})$\end{small}
    \STATE Project $y^{wa}$ to 2D poses $x^{wa}$ with camera parameters 
    \STATE \textcolor{OliveGreen}{// Meta-train on Source data:} 
    \STATE Update  $\mathcal{P}_{t}' = \mathcal{P}_{t} - lr_1 \nabla \mathcal{L}_{MSE}(\mathcal{P}_{t}(x^{or}),y^{or})$
    \FOR{$i\gets 1,.., k$}
        \STATE \textcolor{OliveGreen}{// Meta-test on Weak-augmented data:} 
    \STATE $\mathcal{L}_{weak-test} = {L}_{MSE}(\mathcal{P}_{t}'(x^{wa}),y^{wa})$
    \ENDFOR
    \STATE \textcolor{OliveGreen}{// Meta update on Source and Weak-augmented data:} 
    \STATE \begin{small}$\mathcal{L}_{t+1} = \mathcal{L}_{MSE}(\mathcal{P}_{t}(x^{or}),y^{or})+\gamma {L}_{weak-test}$\end{small}
    \STATE $\mathcal{P}_{t+1} = \mathcal{P}_{t} - lr_2(\partial \mathcal{L}_{t+1} / \partial \mathcal{P}_{t})$
    \STATE Generate strong-augmented 3D poses \begin{small}$y^{sa} = G_{sa}(y^{or})$\end{small}
    \STATE Project $y^{sa}$ to 2D poses $x^{sa}$ with camera parameters 
    \STATE \textcolor{OliveGreen}{// Meta-train on Weak-augmented data:} 
    \STATE Update  $\mathcal{P}_{t+1}' = \mathcal{P}_{t+1} - lr_1 \nabla \mathcal{L}_{MSE}(\mathcal{P}_{t+1}(x^{wa}),y^{wa})$
    \FOR{$i\gets 1,.., k$}
        \STATE \textcolor{OliveGreen}{// Meta-test on Strong-augmented data:} 
    \STATE $\mathcal{L}_{strong-test} = {L}_{MSE}(\mathcal{P}_{t+1}'(x^{sa}),y^{sa})$
    \ENDFOR
    \STATE \textcolor{OliveGreen}{// Meta update on Weak- and Strong-augmented data:} 
    \STATE \begin{small}$\mathcal{L}_{t+2} = \mathcal{L}_{MSE}(\mathcal{P}_{t+1}(x^{wa}),y^{wa})+\gamma {L}_{strong-test}$\end{small}
    \STATE $\mathcal{P}_{t+2} = \mathcal{P}_{t+1} - lr_2(\partial \mathcal{L}_{t+2} / \partial \mathcal{P}_{t+1})$
    
  \end{algorithmic}
\label{alg:meta}
\end{algorithm}

For DG problem in 3D HPE, two principal challenges must be addressed. First, there is the issue of synthesizing data, as detailed in Section \ref{sec:aug}. The second challenge revolves around the effective utilization of synthesized data, a facet often overlooked by current methodologies. Existing DG approaches for 3D HPE \cite{zhang2023poseaug,huang2022dh}, conduct the optimization of the pose estimator based on source and synthesized data independently. Unfortunately, this approach lacks mechanisms for fostering interactions between these two optimization processes, resulting in a deficiency of simulated domain shifts in the optimization trajectory.

In contrast, our proposed model-agnostic strategy incorporates meta-optimization to bolster interactions among source poses, weak-augmented poses, and strong-augmented poses. This process facilitates the learning of domain-invariant knowledge during the update of the pose estimator, as outlined in Algorithm \ref{alg:meta}. The effectiveness of this approach lies in the fact that the objectives in meta-optimization not only aim to minimize losses on source and synthesized poses but also enhance the alignment of optimization directions during training, thus enhancing generalization significantly.

The algorithm can be dissected into two parts: \textcolor{blue}{Lines 1-11} manage interactions between source poses and weakly-augmented poses, while \textcolor{blue}{Lines 12-22} address interactions between weak- and strong-augmented poses. This step-by-step approach is taken due to the substantial domain gap between source poses and strong-augmented poses. Weak-augmented poses serve as a bridge between source poses and strong-augmented poses, alleviating the challenge of directly aligning source and strong-augmented data. By incorporating all three types of poses in the optimization, the pose estimator can effectively generalize across diverse target domains, avoiding overfitting specific pose data types.

\section{Experiments}
\label{sec:experiments}

\subsection{Datasets and Metrics}

In this paper, we evaluate our approach using several widely-used 3D human pose benchmarks, including Human3.6M \cite{ionescu2013human3}, MPI-INF-3DHP \cite{mehta2017monocular}, and 3DPW \cite{von2018recovering}. Moreover, following previous works \cite{zhang2023poseaug,huang2022dh,li2023cee}, we adopt the 16-keypoint human model with Hip joint as the origin.

\noindent\textbf{Human3.6M} is an indoor dataset comprising 3.6 million frames and consisting of 7 subjects denoted as S1, S5, S6, S7, S8, S9, and S11. For the cross-dataset evaluation in Tab. \ref{tab:hp3d}, Tab. \ref{tab:pw3d} and Tab. \ref{tab:backbone-2D}, we use S1, S5, S6, S7, S8 as the source domain. In the cross-scenario evaluation on the entire Human3.6M dataset in Tab. \ref{tab:h36m-full}, S1, S5, S6, S7, S8 are the source domain, while S9 and S11 are the target domain. Mean Per Joint Position Error (MPJPE) and Procrustes-Aligned Mean Per Joint Position Error (PA-MPJPE) are employed as evaluation metrics. For the cross-scenario evaluation on partial Human3.6M, we follow previous works \cite{gholami2022adaptpose,huang2022dh,li2023cee} to define two tasks as shown in Tab. \ref{tab:h36m-part}. One task uses S1 as the source and S5, S6, S7, S8 as the target, while the other task uses S1, S5 as the source and S6, S7, S8 as the target. Both tasks utilize MPJPE as the metric.

\noindent\textbf{MPI-INF-3DHP} (\textbf{3DHP}) is an in-the-wild dataset, and we utilize only its test set for cross-dataset evaluation, as shown in Tab. \ref{tab:hp3d} and Tab. \ref{tab:backbone-2D}, which consists of approximately 3k frames. The results are presented based on three metrics: Percentage of Correct Keypoints (PCK), Area Under the Curve (AUC), and MPJPE.

\noindent\textbf{3DPW} is another in-the-wild dataset featuring more challenging poses and scenes. We utilize it for cross-dataset evaluation, as shown in Tab. \ref{tab:pw3d}. Here PA-MPJPE and MPJPE serve as the evaluation metrics.

\subsection{Implementation Details}

For all the generators and discriminators, we ensure consistency by employing the same fully-connected layers, aligning with the methodology described in \cite{zhang2023poseaug}. In the \emph{data augmentation process}, the learning rate is set to 1e-4 for generators and 2e-4 for discriminators. We set $\alpha_1$ and $\alpha_2$ to 0.50 and 0.35, respectively, while both $\beta_1$ and $\beta_2$ are assigned a value of 4. During the \emph{meta optimization process}, we utilize a learning rate of 1e-4 for $lr_1$ and 5e-4 for $lr_2$. The trade-off parameter $\gamma$ and the inner loop iteration $k$ are both set to 1.

Moreover, we employ the Adam optimizer \cite{kingma2014adam} for data augmentation and the AdamW optimizer \cite{loshchilov2018decoupled} for meta optimization. Our experiments are conducted with a batch size of 1024 over 60 epochs. We initialize the pose estimator with a warm-up phase lasting two epochs for supervised learning on source data. From the third epoch onwards, data augmentation and meta-optimization begin. 

\subsection{Quantitative Results}

\vspace{-5pt}
\begin{table}[!ht]
    \scriptsize
    \centering
    \caption{Cross-dataset evaluation on 3DHP dataset.}
    \resizebox{1.0\linewidth}{!}{%
    \begin{tabular}{c|c|c|c|c|c}
          \toprule
          Method  & Venue & DG & PCK $\uparrow$ & AUC $\uparrow$ & MPJPE $\downarrow$  \\
         \hline 
         VPose (1-frame) \cite{pavllo20193d} & CVPR'19 & $\times$ & 80.9 & 42.5 & 102.3\\
         \hline
         EvoSkeleton \cite{li2020cascaded} & CVPR'20 & $\checkmark$ & 81.2 & 46.1 & 99.7\\
         RepNet \cite{wandt2019repnet} & CVPR'19 & $\checkmark$ & 81.8 & 54.8 & 92.5\\
         PoseAug \cite{zhang2023poseaug} & TPAMI'23 & $\checkmark$ & 88.6 & 57.3 & 73.0\\
         DH-AUG \cite{huang2022dh} & ECCV'22 & $\checkmark$ & 89.5 & 57.9 & 71.2\\
         PoseGU \cite{guan2023posegu} & CVIU'23 & $\checkmark$ & 86.3 & 57.2 & 75.0\\
         CEE-Net \cite{li2023cee} & AAAI'23 & $\checkmark$ & 89.9 & 58.2 & 69.7 \\
         \hline
         Ours &  &  $\checkmark$ &  \textbf{92.9} &  \textbf{60.7} &  \textbf{63.1}\\

         \bottomrule
    \end{tabular}%
    }
    \vspace{-7pt}
\label{tab:hp3d}
\end{table}

\begin{table}[!ht]
    \scriptsize
    \centering
    \caption{Cross-dataset evaluation on 3DPW dataset.}
    \resizebox{1.0\linewidth}{!}{%
    \begin{tabular}{c|c|c|c|c}
          \toprule
          Method & Venue & DG & PA-MPJPE $\downarrow$ & MPJPE $\downarrow$  \\
         \hline 
         VPose (1-frame) \cite{pavllo20193d}& CVPR'19 & $\times$ & 94.6 & 125.7\\
         \hline
         VIBE \cite{kocabas2020vibe} & CVPR'20 & $\checkmark$ & 82.3 & 122.5 \\
         PoseAug \cite{zhang2023poseaug} & TPAMI'23 & $\checkmark$ & 81.6 & 119.0\\
         DH-AUG \cite{huang2022dh}  & ECCV'22 & $\checkmark$ & 79.3 & 112.8\\
         PoseGU \cite{guan2023posegu} & CVIU'23 & $\checkmark$ & 92.3 & - \\
         CEE-Net \cite{li2023cee} & AAAI'23 & $\checkmark$ & 76.8 & - \\
         \hline
         Ours & &  $\checkmark$ &  \textbf{73.2} &  \textbf{106.6} \\

         \bottomrule
    \end{tabular}%
    }
    \vspace{-7pt}
\label{tab:pw3d}
\end{table}

\begin{table}[!ht]
    \scriptsize
    \centering
    \caption{Cross-scenario evaluation on Entire Human3.6M dataset. S1,S5,S6,S7,S8 are the source and S9,S11 are the target.}
    \resizebox{1.0\linewidth}{!}{%
    \begin{tabular}{c|c|c|c|c}
          \toprule
          Method & Venue & DG & MPJPE $\downarrow$ & PA-MPJPE $\downarrow$  \\
         \hline 
         VPose (1-frame) \cite{pavllo20193d} & CVPR'19 & $\times$ & 52.7 & 40.9\\
         \hline
         EvoSkeleton \cite{li2020cascaded} & CVPR'20 & $\checkmark$ & 50.9 & 38.0 \\
         PoseAug \cite{zhang2023poseaug} & TPAMI'23 & $\checkmark$ & 50.2 & 39.1\\
         DH-AUG \cite{huang2022dh}  & ECCV'22 & $\checkmark$ & 49.8 & 38.3\\
         CEE-Net \cite{li2023cee} & AAAI'23 & $\checkmark$ & 47.3 & 36.8 \\
         \hline
         Ours & &  $\checkmark$ &  \textbf{44.4} &  \textbf{34.6} \\

         \bottomrule
    \end{tabular}%
    }
\label{tab:h36m-full}
\end{table}

\begin{table}[!ht]
    \scriptsize
    \centering
    \caption{Cross-scenario evaluation on Partial Human3.6M dataset. For the task "S1", S1 is the source and S5, S6, S7, S8 are the target. For the task "S1+S5", S1 and S5 are the source, and S6, S7, S8 are the target. MPJPE ($\downarrow$) is used for evaluation.}
    \resizebox{1.0\linewidth}{!}{%
    \begin{tabular}{c|c|c|c|c}
          \toprule
          Method & Venue & DG & S1 & S1+S5  \\
         \hline 
         VPose (1-frame) \cite{pavllo20193d} & CVPR'19 & $\times$ & 65.2 & 57.9\\
         \hline
         EvoSkeleton \cite{li2020cascaded} & CVPR'20 & $\checkmark$ & 61.5 & 54.6\\
         PoseAug \cite{zhang2023poseaug} & TPAMI'23 & $\checkmark$ & 56.7 & 51.3\\
         DH-AUG \cite{huang2022dh} & ECCV'22 & $\checkmark$ & 52.2 & 47.0\\
         CEE-Net \cite{li2023cee} & AAAI'23 & $\checkmark$ & 51.9 & 46.7 \\
         \hline
         Ours & &  $\checkmark$ &  \textbf{50.3} &  \textbf{45.4} \\

         \bottomrule
    \end{tabular}%
    }
    \vspace{-5pt}
\label{tab:h36m-part}
\end{table}

\vspace{-5pt}
\begin{table}[!ht]
    \scriptsize
    \centering
    \caption{Cross-dataset evaluation with MPJPE ($\downarrow$) on 3DHP with varied 2D predictions and 2D-to-3D backbones (1-frame). }
    \resizebox{1.0\linewidth}{!}{%
    \begin{tabular}{c|c|c|c|c|c}
          \toprule
          Method  & DG & DET \cite{detectron2018} & CPN \cite{chen2018cascaded} & HR \cite{wang2020deep} & GT  \\
         \hline
         SemGCN \cite{zhao2019semgcn} & $\times$ & 101.9 & 98.7 & 95.6 & 97.4 \\
         SemGCN + PoseAug \cite{zhang2023poseaug}& $\checkmark$ & 89.9 & 89.3 & 89.1 & 86.1\\
         SemGCN + CEE-generator \cite{li2023cee} & $\checkmark$ & 83.6 & 82.8 & 82.4 & 81.3\\
         SemGCN + DH-AUG \cite{huang2022dh} & $\checkmark$ & 79.7 & 76.7 & 73.0 & 71.3\\
         SemGCN + Ours & $\checkmark$ & \textbf{76.5} &  \textbf{74.1} & \textbf{70.7} & \textbf{68.9}\\
         \hline 
         VPose \cite{pavllo20193d} & $\times$ & 92.6 & 89.8 & 85.6 & 86.6 \\
         VPose + PoseAug \cite{zhang2023poseaug}& $\checkmark$ & 78.3 & 78.4 & 73.2 & 73.0\\
         VPose + CEE-generator \cite{li2023cee} & $\checkmark$ & 75.6 & 75.2 & 71.2 & 71.4\\
         VPose + DH-AUG \cite{huang2022dh} & $\checkmark$ & 76.7 & 74.8 & 71.1 & 71.2\\
         VPose + Ours & $\checkmark$ & \textbf{72.4} & \textbf{70.9} & \textbf{62.4} & \textbf{63.1}\\
         \hline
         PoseFormer \cite{zheng20213d} & $\times$ & 91.9 & 89.2 & 84.2 & 85.7\\
         PoseFormer + PoseAug \cite{zhang2023poseaug}& $\checkmark$ & 77.7 & 77.5 & 72.1 & 72.3\\
         PoseFormer + CEE-generator \cite{li2023cee} & $\checkmark$ & - & - & - & - \\
         PoseFormer + DH-AUG \cite{huang2022dh} & $\checkmark$ & 75.6 & 74.8 & 71.6 & 72.0\\
         PoseFormer + Ours & $\checkmark$ & \textbf{72.2} & \textbf{70.5} & \textbf{62.8} & \textbf{63.4}\\
         \hline
         MixSTE \cite{zhang2022mixste} & $\times$ & 90.6 & 87.4 & 82.0 & 84.0\\
         MixSTE + PoseAug \cite{zhang2023poseaug}& $\checkmark$ & 76.1 & 76.3 & 71.7 & 71.6\\
         MixSTE + CEE-generator \cite{li2023cee} & $\checkmark$ & - & - & - & - \\
         MixSTE + DH-AUG \cite{huang2022dh} & $\checkmark$ & 74.8 & 74.4 & 70.9 & 70.7\\
         MixSTE + Ours & $\checkmark$ & \textbf{70.5} & \textbf{68.2} & \textbf{60.4} & \textbf{61.0}\\
        

         \bottomrule
    \end{tabular}%
    }
    \vspace{-15pt}
\label{tab:backbone-2D}
\end{table}

\begin{figure*}[!ht]
  \centering
   \includegraphics[width=1.00\linewidth]{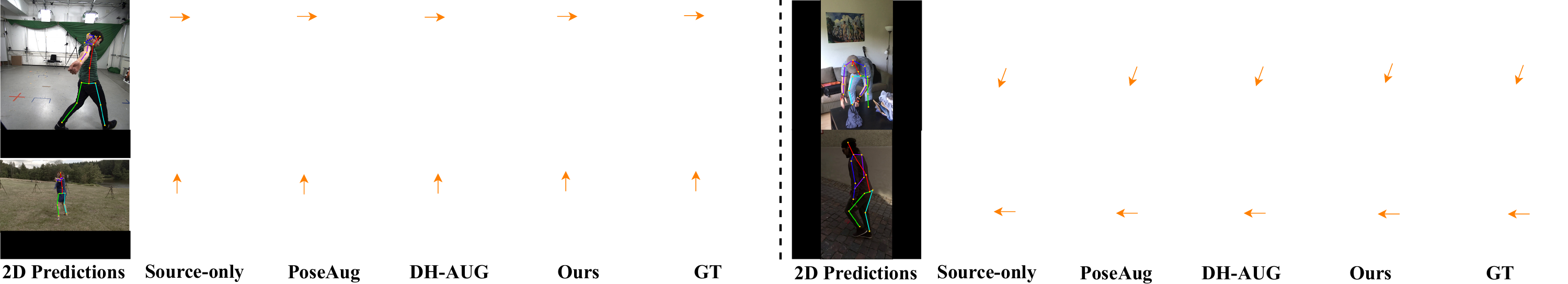}

   \caption{ Qualitative results on Cross-dataset evaluation. Left is 3DHP dataset, and right is 3DPW dataset. 
   } %
   \vspace{-5pt}
   \label{fig:xdata}
\end{figure*}

\begin{figure*}[t]
  \centering
   \includegraphics[width=1.00\linewidth]{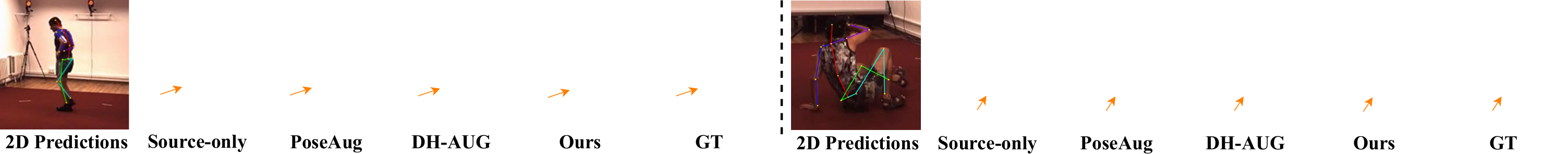}

   \caption{Results on Cross-scenario evaluation. Left is for task S1,S5,S6,S7,S8 $\rightarrow$ S9,S11, and right is for task S1,S5 $\rightarrow$ S6,S7,S8. 
   } %
   \vspace{-10pt}
   \label{fig:xscene}
\end{figure*}

\noindent\textbf{Cross-dataset evaluation results.} In cross-dataset evaluations, source and target come from different datasets. Following identical paradigms from existing methods \cite{zhang2023poseaug,huang2022dh,li2023cee}, we use ground truth 2D keypoints as input, single-frame VPose \cite{pavllo20193d} as the lifting backbone, and Human3.6M as the source dataset. Our method demonstrates notable performance in all metrics, as presented in Tab. \ref{tab:hp3d} and Tab. \ref{tab:pw3d}. Notably, our approach outperforms CEE-Net by $3.0\%$ in PCK and $2.5\%$ in AUC, and reduces MPJPE by $6.6$mm in the 3DHP task. In the case of 3DPW, our method shows an improvement of $3.6$mm in PA-MPJPE compared to CEE-Net \cite{li2023cee}. While CEE-Net \cite{li2023cee} and PoseGU \cite{guan2023posegu} do not disclose their codes or report their results on MPJPE, it is evident that our method surpasses DH-AUG \cite{huang2022dh} by $6.2$mm.

\noindent\textbf{Cross-scenario evaluation results.} In cross-scenario evaluations, source and target come from different subsets of the same dataset. Maintaining consistency with previous works, we utilize ground truth 2D keypoints as input and single-frame VPose \cite{pavllo20193d} as the 2D-to-3D lifting network. For the situation of using Entire Human3.6M in Tab. \ref{tab:h36m-full}, our method demonstrates superior performance compared to CEE-Net \cite{li2023cee} with a $2.9$mm reduction in MPJPE and a $2.2$mm improvement in PA-MPJPE. In the case of using partial Human3.6M in Tab. \ref{tab:h36m-part}, our approach surpasses CEE-Net \cite{li2023cee} by $1.6$mm in the S1 task and $1.3$mm in the S1+S5 task based on the MPJPE metric.

\noindent\textbf{Results with varied 2D predictions and backbones.} The results presented in Tables \ref{tab:hp3d} to \ref{tab:h36m-part} are confined to the usage of ground truth 2D input and single-frame VPose backbone, which may raise concerns about the universality of the proposed method. To address this concern, we assess the performance of our approach with various 2D predictions such as DET \cite{detectron2018}, CPN \cite{chen2018cascaded}, HRNet \cite{wang2020deep}, and diverse lifting backbones including SemGCN \cite{zhao2019semgcn}, PoseFormer \cite{zheng20213d}, and MixSTE \cite{zhang2022mixste}, as displayed in Table \ref{tab:backbone-2D}. In this evaluation, 3DHP serves as the dataset, and MPJPE is the metric used. Notably, all the listed backbones are single-frame versions. As CEE-Net \cite{li2023cee} only provides results for its generation part, CEE-generator, and does not offer open-source code, we have included partial results of CEE-generator. From Table \ref{tab:backbone-2D}, it is evident that our method surpasses all the existing methods, demonstrating the robustness of our framework across various settings. 

\subsection{Qualitative Results}

Fig. \ref{fig:xdata} shows qualitative results on cross-dataset evaluation (3DHP on the left side and 3DPW on the right side), while Fig. \ref{fig:xscene} displays qualitative results on cross-scenario evaluation (S1,S5,S6,S7,S8 $\rightarrow$ S9,S11 on the left side and S1,S5 $\rightarrow$ S6,S7,S8 on the right side). HRNet \cite{wang2020deep} is applied as the 2D pose estimator and VPose \cite{pavllo20193d} is the 2D-to-3D lifting backbone. We use \textbf{Source-only}, \textbf{PoseAug} \cite{zhang2023poseaug}, \textbf{DH-AUG} \cite{huang2022dh}, \textbf{Ours}, and \textbf{Ground Truth (GT)} for qualitative comparison. \ul{Because CEE-Net does not provide source codes or pretrained models, we cannot generate visual examples from it.} It is evident that our method outperforms other baselines significantly.

\subsection{Ablation Study}

\noindent\textbf{Ablation study on the overall framework.} The ablation study is conducted to illustrate the functions of three proposed modules: differential generation (\textbf{DiffGen}) in Sec. \ref{sec:aug}, differential discrimination (\textbf{DiffDis}) in Sec. \ref{sec:aug}, and meta optimization (\textbf{MetaOpt}) in Sec. \ref{sec:meta}. In Tab. \ref{tab:ab1-vpose}, we apply ground truth as 2D predictions and VPose as the backbone. The absence of DiffGen leads to a decrease in PCK and AUC by $2.4\%$ and $2.1\%$ respectively, accompanied by a 5.9mm increase in MPJPE on 3DHP, while it increases PA-MPJPE and MPJPE by 3.1mm and 5.2mm separately on 3DPW.  Similarly, the exclusion of DiffDis results in a decrease of $1.7\%$ in both PCK and AUC, with a corresponding 4.1mm increase in MPJPE on 3DHP. As for 3DPW, the removal causes a degradation of 1.9mm in PA-MPJPE and 3.0mm in MPJPE. Besides, the removal of MetaOpt leads to a decline in PCK and AUC by $1.2\%$ and $0.8\%$ respectively, along with a 2.4mm increase in MPJPE on 3DHP, and an increase of 1.3mm and 1.8mm in PA-MPJPE and MPJPE separately. These results show that each module plays a critical role in obtaining better generalization. 

\begin{table}[!ht]
    \scriptsize
    \centering
    \caption{Overall framework ablation study on 3DHP and 3DPW}
    \resizebox{1.0\linewidth}{!}{%
    \begin{tabular}{c|c|c|c|c|c}
         \toprule
          & \multicolumn{3}{c|}{3DHP} & \multicolumn{2}{c}{3DPW} \\
         \hline 
         Method  & PCK $\uparrow$ & AUC $\uparrow$ & MPJPE $\downarrow$ & PA-MPJPE $\downarrow$ & MPJPE $\downarrow$  \\ 
         \hline
         Ours w/o DiffGen & 90.5 & 58.6 & 69.0 & 76.3 & 111.8\\
         Ours w/o DiffDis & 91.2 & 59.0 & 67.3 & 75.1 & 109.6\\
         Ours w/o MetaOpt & 91.7 & 59.9 & 65.5 & 74.5 & 108.4\\
         Ours & \textbf{92.9} &  \textbf{60.7} &  \textbf{63.1} & \textbf{73.2} & \textbf{106.6}\\

         \bottomrule
    \end{tabular}%
    }
\label{tab:ab1-vpose}
\end{table}

\noindent\textbf{Ablation study on the generators.} There exist two generators in our framework, and each with two pair groups. In this part, we discuss the functions of proximate pairs in weak augmentor (\textbf{W-PP}), proximate pairs in strong augmentor (\textbf{S-PP}), one-state gap pairs in weak augmentor (\textbf{W-OG}), and one-state gap pairs in strong augmentor (\textbf{S-OG}). 

\begin{table}[!ht]
    \scriptsize
    \centering
    \caption{Generators' ablation study on 3DHP and 3DPW}
    \resizebox{1.0\linewidth}{!}{%
    \begin{tabular}{c|c|c|c|c|c}
         \toprule
          & \multicolumn{3}{c|}{3DHP} & \multicolumn{2}{c}{3DPW} \\
         \hline 
         Method  & PCK $\uparrow$ & AUC $\uparrow$ & MPJPE $\downarrow$ & PA-MPJPE $\downarrow$ & MPJPE $\downarrow$  \\ 
         \hline
         Ours w/o W-PP & 88.3 & 57.5 & 72.6 & 81.7 & 118.8\\
         Ours w/o S-PP & 90.8 & 58.2 & 71.3 & 78.1 & 111.0\\
         Ours w/o W-OG & 92.1 & 59.6 & 65.8 & 74.7 & 108.7\\
         Ours w/o S-OG & 91.4 & 58.9 & 68.2 & 75.4 & 109.5\\
         Ours & \textbf{92.9} &  \textbf{60.7} &  \textbf{63.1} & \textbf{73.2} & \textbf{106.6}\\ 
            
         \bottomrule
    \end{tabular}%
    }
\label{tab:ab3-vpose}
\end{table}

In Table \ref{tab:ab3-vpose}, excluding W-PP or S-PP leads to a significant decline in PCK by $4.6\%$ and $2.1\%$, and in AUC by $3.2\%$ and $2.5\%$ respectively, accompanied by a notable increase of 9.5mm and 8.2mm in MPJPE separately on 3DHP. 
These results emphasize the critical role of maintaining similarity in proximate pairs for both weak and strong augmentors, serving as the fundamental basis for generating effective and reasonable synthesized poses. Moreover, the absence of W-OG leads to a decline in PCK and AUC by $0.8\%$ and $1.1\%$ respectively, with a corresponding 1.7mm increase in the MPJPE on 3DHP. The removal of S-OG results in a decrease in PCK and AUC scores by $1.5\%$ and $1.8\%$ respectively, along with a 3.4mm increase in MPJPE on 3DHP. 
These results highlight the significance of maintaining differentiation between the weak augmentor and the strong augmentor during the generation process, where enlarging dissimilarity in S-OG is more important in discriminating these two generators. 

\noindent\textbf{Ablation study on the number of augmentors.} Comparisons were conducted between our dual-augmentor framework and single-augmentor frameworks. Alongside our proposed framework, two single-augmentor frameworks were considered in the ablation study, utilizing either the weak-augmentor (\textbf{WA}) or the strong-augmentor (\textbf{SA}). The discrimination and meta-optimization processes exclusively involved source poses and one category of synthesized poses. The results, using ground truth as 2D predictions and VPose as the backbone, are presented in Table \ref{tab:ab-one}.

\begin{table}[!ht]
    \scriptsize
    \centering
    \caption{Ablation study of number of augmentors on 3DHP and 3DPW}
    \resizebox{1.0\linewidth}{!}{%
    \begin{tabular}{c|c|c|c|c|c}
         \toprule
          & \multicolumn{3}{c|}{3DHP} & \multicolumn{2}{c}{3DPW} \\
         \hline 
         Method  & PCK $\uparrow$ & AUC $\uparrow$ & MPJPE $\downarrow$ & PA-MPJPE $\downarrow$ & MPJPE $\downarrow$  \\ 
         \hline
         WA & 87.3 & 56.0 & 74.5 & 80.5 & 117.7\\
         SA & 89.8 & 57.8 & 71.0 & 79.1 & 111.4\\
         Ours & \textbf{92.9} &  \textbf{60.7} &  \textbf{63.1} & \textbf{73.2} & \textbf{106.6}\\ 
            
         \bottomrule
    \end{tabular}%
    }
    \vspace{-10pt}
\label{tab:ab-one}
\end{table}

From Table \ref{tab:ab-one}, it is evident that our proposed framework surpasses both WA and SA significantly, underscoring the superiority of employing two augmentors over a single augmentor in addressing DG for 3D HPE. Furthermore, SA outperforms WA, emphasizing the greater significance of exploring out-of-source distributions compared to retaining source-relevant knowledge in cross-dataset tasks.
\section{Conclusion}
\label{sec:conclusion}

In this paper, we propose a novel dual-augmentor framework designed to enhance domain generalization in 3D human pose estimation. Our framework addresses the critical aspects of data augmentation and the effective utilization of synthesized data. To achieve this, we implement distinctive strategies for the weak and strong generators, ensuring the preservation of source-specific information while simultaneously exploring out-of-source distributions. Moreover, we incorporate meta-optimization techniques to facilitate enhanced interaction among source data, weak-augmented data, and strong-augmented data, thereby simulating domain shifts in the training of pose estimator and fostering the acquisition of domain-invariant knowledge. Extensive experimentation and comprehensive analysis conducted across multiple datasets demonstrate the superior performance of our proposed approach over existing state-of-the-art methods.

\noindent\textbf{Acknowledgements} This material is based upon work supported by the National Science Foundation under Grant CNS-1910844.

{
    \small
    \bibliographystyle{ieeenat_fullname}
    \bibliography{main}
}

\clearpage
\setcounter{page}{1}
\maketitlesupplementary

\section{Overview}

The supplementary material is organized into the following sections:

\begin{itemize}
    \item Section \ref{sec:qua}: Additional qualitative results on on cross-dataset evaluation.
    \item Section \ref{sec:ab-f}: Additional ablation study on overall framework.
    \item Section \ref{sec:ab-g}: Additional ablation of generation process.
    \item Section \ref{sec:ab-a}: Additional ablation of the number of augmentors.
    \item Section \ref{sec:param}: Hyperparameters' analysis.
    \item Section \ref{sec:multi}: Experiments on the multi-frame setting.
    \item Section \ref{sec:vis}: Visualizations of distributions between source data and synthesized data.
    \item Section \ref{sec:cmp}: Visualizations of weak-augmented and strong-augmented poses.
    \item Section \ref{sec:laplacian}: Details of Laplacian weighted similarity.
   
\end{itemize}

\section{Extra Qualitative Results on Cross-dataset Evaluation}
\label{sec:qua}

\begin{figure*}[!ht]
  \centering
   \includegraphics[width=1.00\linewidth]{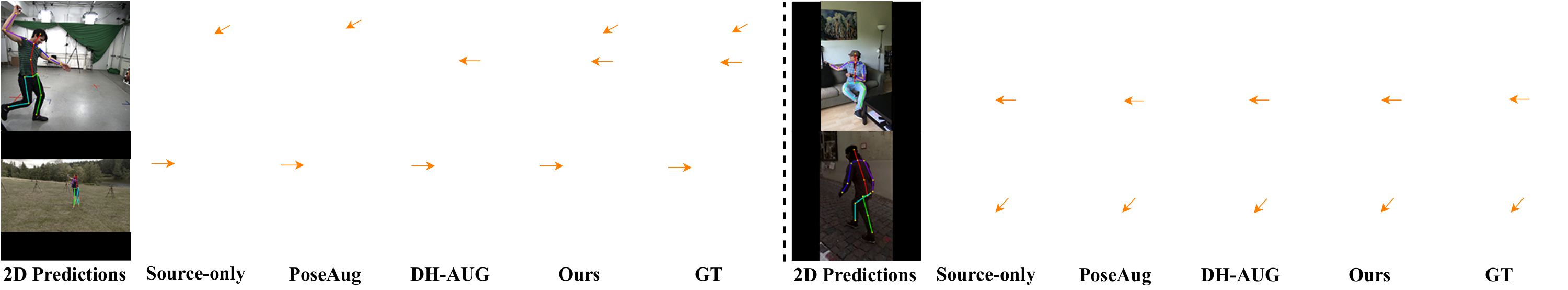}

   \caption{ Extra qualitative results on cross-dataset evaluation. Left is the 3DHP dataset, and right is the 3DPW dataset. 
   } %
   \label{fig:xdata_sm}
\end{figure*}

Fig. \ref{fig:xdata_sm} shows extra qualitative results on cross-dataset evaluation (3DHP on the left side and 3DPW on the right side). HRNet \cite{wang2020deep} is applied as the 2D pose estimator and VPose \cite{pavllo20193d} is the 2D-to-3D lifting backbone. We use \textbf{Source-only}, \textbf{PoseAug} \cite{zhang2023poseaug}, \textbf{DH-AUG} \cite{huang2022dh}, \textbf{Ours}, and \textbf{Ground Truth (GT)} for qualitative comparison. It is evident that our method outperforms other baselines significantly. 

\section{Extra Ablation Study on the Overall Framework}
\label{sec:ab-f}
\begin{table}[!ht]
    \scriptsize
    \centering
    \caption{Ablation study on the overall framework with varied 2D pose predictions and 2D-to-3D lifting backbones.}
    \resizebox{1.0\linewidth}{!}{%
    \begin{tabular}{c|c|c|c|c|c}
         \toprule
         Method & 2D-to-3D Lifting Backbones  & \multicolumn{4}{c}{2D Pose Predictions} \\
         \hline
         &  & DET & CPN & HR & GT  \\
         \hline
         Ours w/o DiffGen & SemGCN & 80.1 & 78.8 & 74.0 & 71.8\\
         Ours w/o DiffDis & SemGCN & 79.4 & 77.2 & 73.5 & 70.9\\
         Ours w/o MetaOpt & SemGCN & 77.9 & 75.7 & 72.4 & 70.3\\
         Ours & SemGCN & \textbf{76.5} &  \textbf{74.1} & \textbf{70.7} & \textbf{68.9}\\
         \hline 
         Ours w/o DiffGen &  VPose & 77.1 & 76.7 & 68.6 & 69.0\\
         Ours w/o DiffDis &  VPose & 76.7 & 74.8 & 67.0 & 67.3\\
         Ours w/o MetaOpt &  VPose & 74.5 & 72.4 & 64.9 & 65.5\\
         Ours &  VPose & \textbf{72.4} & \textbf{70.9} & \textbf{62.4} & \textbf{63.1}\\
         \hline
         Ours w/o DiffGen & PoseFormer & 76.7 & 75.9 & 68.8 & 69.5\\
         Ours w/o DiffDis & PoseFormer & 76.1 & 75.0 & 67.2 & 67.7 \\
         Ours w/o MetaOpt & PoseFormer & 74.5 & 72.3 & 65.3 & 66.1\\
         Ours & PoseFormer & \textbf{72.2} & \textbf{70.5} & \textbf{62.8} & \textbf{63.4}\\
         \hline
         Ours w/o DiffGen & MixSTE & 75.4 & 74.9 & 71.8 & 71.2\\
         Ours w/o DiffDis & MixSTE & 74.7 & 73.0 & 66.6 & 66.4\\
         Ours w/o MetaOpt & MixSTE & 72.8 & 71.3 & 63.4 & 63.5 \\
         Ours &  MixSTE & \textbf{70.5} & \textbf{68.2} & \textbf{60.4} & \textbf{61.0}\\

         \bottomrule
    \end{tabular}%
    }
\label{tab:ab2-backbone}
\end{table}

Tab. \ref{tab:ab2-backbone} offers valuable insights into the impact of using different 2D pose detectors like \textbf{DET} \cite{detectron2018}, \textbf{CPN} \cite{chen2018cascaded}, HRNet (\textbf{HR}) \cite{wang2020deep} and Ground Truth (\textbf{GT}) for 2D poses predictions, and 2D-to-3D backbones \textbf{SemGCN} \cite{zhao2019semgcn}, \textbf{VPose} \cite{pavllo20193d}, \textbf{PoseFormer} \cite{zheng20213d}, \textbf{MixSTE} \cite{zhang2022mixste} on the generalization for the ablation on the overall framework. Here we use MPJPE values on the 3DHP \cite{von2018recovering} dataset for comparisons. For instance, with GT as 2D predictions and PoseFormer as the backbone, the absence of differential generation (DiffGen) leads to a 6.1mm increase, while excluding differential discrimination (DiffDis) results in a corresponding 4.3mm increase. As for the removal of meta optimization (MetaOpt), it causes a degradation of 2.7mm. The results highlight the integral role of each module in enhancing the overall framework's efficacy across diverse 2D predictions and backbones, affirming our method's robustness and versatility in improving DG for 3D HPE performance.

\section{Extra Ablation Study on the Generation Process}
\label{sec:ab-g}
\begin{table}[!ht]
    \scriptsize
    \centering
    \caption{Ablation study on the generation process with varied 2D pose predictions and 2D-to-3D lifting backbones.}
    \resizebox{1.0\linewidth}{!}{%
    \begin{tabular}{c|c|c|c|c|c}
          \toprule
           Method & 2D-to-3D Lifting Backbones  & \multicolumn{4}{c}{2D Pose Predictions} \\
           \hline
          &  & DET & CPN & HR & GT  \\
         \hline 
         Ours w/o W-PP & SemGCN & 82.0 & 80.4 & 74.7 & 72.9\\
         Ours w/o S-PP & SemGCN & 80.3 & 79.2 & 73.9 & 71.7\\
         Ours w/o W-OG & SemGCN & 77.9 & 75.6 & 72.5 & 69.8\\
         Ours w/o S-OG & SemGCN & 79.8 & 78.7 & 74.3 & 71.1 \\
         Ours &  SemGCN & \textbf{76.5} &  \textbf{74.1} & \textbf{70.7} & \textbf{68.9}\\
         \hline
         Ours w/o W-PP & VPose & 79.5 & 78.8 & 73.4 & 72.6\\
         Ours w/o S-PP & VPose & 77.1 & 76.9 & 72.2 & 71.3\\
         Ours w/o W-OG & VPose & 73.7 & 73.1 & 65.7 & 65.8\\
         Ours w/o S-OG & VPose & 76.2 & 75.6 & 68.4 & 68.2\\
         Ours &  VPose & \textbf{72.4} & \textbf{70.9} & \textbf{62.4} & \textbf{63.1}\\
         \hline
         Ours w/o W-PP & PoseFormer & 78.9 & 78.2 & 73.7 & 73.2\\
         Ours w/o S-PP & PoseFormer & 76.7 & 76.4 & 73.0 & 72.1\\
         Ours w/o W-OG & PoseFormer & 73.3 & 72.9 & 65.6 & 66.3\\
         Ours w/o S-OG & PoseFormer & 76.0 & 75.6 & 68.8 & 68.5\\
         Ours & PoseFormer & \textbf{72.2} & \textbf{70.5} & \textbf{62.8} & \textbf{63.4}\\
         \hline
         Ours w/o W-PP & MixSTE & 78.3 & 77.1 & 72.5 & 72.4\\
         Ours w/o S-PP & MixSTE & 75.9 & 74.2 & 66.3 & 66.8\\
         Ours w/o W-OG & MixSTE & 72.0 & 70.6 & 62.6 & 63.3\\
         Ours w/o S-OG & MixSTE & 74.8 & 73.4 & 64.5 & 64.7\\
         Ours &  MixSTE & \textbf{70.5} & \textbf{68.2} & \textbf{60.4} & \textbf{61.0}\\

         \bottomrule
    \end{tabular}%
    }
\label{tab:ab4-gene}
\end{table}

Tab. \ref{tab:ab4-gene} provides observations regarding the influence of various 2D predictions and 2D-to-3D backbones on the generalization in the context of the generator ablation study. Here we use MPJPE values on the 3DHP dataset \cite{von2018recovering} for comparisons. For instance, with GT as 2D predictions and PoseFormer as the backbone, the absence of W-PP leads to a 9.8mm increase, while excluding S-PP results in a corresponding 8.7mm increase. As for the removal of W-OG and S-OG, they cause a degradation of 2.9mm and 5.1 respectively. The results underscore the crucial contribution of each module in augmenting the effectiveness of the generation process across various 2D predictions and backbone models. This affirms the robustness and versatility of our method in enhancing the performance of DG for 3D HPE.

\section{Extra Ablation Study on the Number of Augmentors}
\label{sec:ab-a}

\begin{table}[!ht]
    \scriptsize
    \centering
    \caption{Ablation study on the number of augmentors with varied 2D pose predictions and 2D-to-3D lifting backbones.}
    \resizebox{1.0\linewidth}{!}{%
    \begin{tabular}{c|c|c|c|c|c}
           \toprule
           Method & 2D-to-3D Lifting Backbones  & \multicolumn{4}{c}{2D Pose Predictions} \\
           \hline
          &  & DET & CPN & HR & GT  \\
         \hline 
         WA & SemGCN & 88.4 & 87.7 & 87.3 & 85.3\\
         SA & SemGCN & 81.7 & 80.5 & 77.2 & 76.9\\
         Ours & SemGCN & \textbf{76.5} &  \textbf{74.1} & \textbf{70.7} & \textbf{68.9}\\
         \hline 
         WA & VPose & 79.5 & 77.6 & 75.0 & 74.5\\
         SA & VPose & 76.1 & 75.9 & 72.7 & 71.0\\
         Ours &  VPose & \textbf{72.4} & \textbf{70.9} & \textbf{62.4} & \textbf{63.1}\\
         \hline
         WA & PoseFormer & 80.8 & 78.2 & 75.4 & 75.1\\
         SA & PoseFormer & 75.0 & 74.6 & 70.5 & 70.7\\
         Ours & PoseFormer & \textbf{72.2} & \textbf{70.5} & \textbf{62.8} & \textbf{63.4}\\
         \hline
         WA & MixSTE & 77.3 & 78.0 & 73.5 & 73.2\\
         SA & MixSTE & 74.1 & 74.4 & 70.2 & 70.3\\
         Ours &  MixSTE & \textbf{70.5} & \textbf{68.2} & \textbf{60.4} & \textbf{61.0}\\

         \bottomrule
    \end{tabular}%
    }
\label{tab:ab4-aug}
\end{table}

In Tab. \ref{tab:ab4-aug}, we compare our dual-augmentor framework with single-augmentor frameworks across different 2D predictions and 2D-to-3D backbones on the generalization for 3D HPE. Here we use MPJPE values on 3DHP \cite{von2018recovering} for comparisons. For instance, with GT as 2D predictions and PoseFormer as the backbone, Ours outperforms SA by 7.3mm, and it surpasses WA by 11.7mm. The results highlight the importance of dual-augmentor frameworks across diverse 2D predictions and backbones, affirming our method's robustness and versatility in improving DG for 3D HPE.

\section{Hyperparameter Analysis}
\label{sec:param}

\begin{table}[!ht]
    \scriptsize
    \centering
    \caption{Parameter analysis of $\alpha_1$ on 3DHP and 3DPW}
    \resizebox{1.0\linewidth}{!}{%
    \begin{tabular}{c|c|c|c|c|c}
         \toprule
          & \multicolumn{3}{c|}{3DHP} & \multicolumn{2}{c}{3DPW} \\
         \hline 
         $\alpha_1$  & PCK $\uparrow$ & AUC $\uparrow$ & MPJPE $\downarrow$ & PA-MPJPE $\downarrow$ & MPJPE $\downarrow$  \\ 
         \hline
         0.35 & 92.2 & 60.0 & 63.8 & 74.5 & 107.9\\
         0.40 & 92.7 & 60.4 & 63.5 & 74.1 & 107.5\\
         0.45 & 92.9 & 60.7 & 63.1 & 73.5 & 106.9\\
         \hline
         0.50 & 92.9 & 60.7 &  63.1 & 73.2 & 106.6\\
         \hline
         0.55 & 92.9 & 60.5 & 63.6 & 73.8 & 107.0 \\
         0.60 & 92.5 & 60.4 & 63.8 & 74.0 & 107.3\\
         0.65 & 92.1 & 60.1 & 63.9 & 74.4 & 107.7\\
            
         \bottomrule
    \end{tabular}%
    }
\label{tab:ab3-alpha1}
\end{table}

\begin{table}[!ht]
    \scriptsize
    \centering
    \caption{Parameter analysis of $\alpha_2$ on 3DHP and 3DPW}
    \resizebox{1.0\linewidth}{!}{%
    \begin{tabular}{c|c|c|c|c|c}
         \toprule
          & \multicolumn{3}{c|}{3DHP} & \multicolumn{2}{c}{3DPW} \\
         \hline 
         $\alpha_2$  & PCK $\uparrow$ & AUC $\uparrow$ & MPJPE $\downarrow$ & PA-MPJPE $\downarrow$ & MPJPE $\downarrow$  \\ 
         \hline
         0.20 & 92.5 & 60.2 & 63.8 & 74.1 & 107.5\\
         0.25 & 92.7 & 60.5 & 63.5 & 73.5 & 106.9\\
         0.30 & 92.9 & 60.7 &  63.1 & 73.2 & 106.6\\
         \hline
         0.35 & 92.9 & 60.7 &  63.1 & 73.2 & 106.6\\
         \hline
         0.40 & 92.9 & 60.7 &  63.1 & 73.2 & 106.6\\
         0.45 & 92.7 & 60.4 & 63.4 & 73.4 & 106.8\\
         0.50 & 92.4 & 60.0 & 63.6 & 73.8 & 107.0\\
            
         \bottomrule
    \end{tabular}%
    }
\label{tab:ab3-alpha2}
\end{table}

\begin{table}[!ht]
    \scriptsize
    \centering
    \caption{Parameter analysis of $\beta$ on 3DHP and 3DPW}
    \resizebox{1.0\linewidth}{!}{%
    \begin{tabular}{c|c|c|c|c|c}
         \toprule
          & \multicolumn{3}{c|}{3DHP} & \multicolumn{2}{c}{3DPW} \\
         \hline 
         $\beta$  & PCK $\uparrow$ & AUC $\uparrow$ & MPJPE $\downarrow$ & PA-MPJPE $\downarrow$ & MPJPE $\downarrow$  \\ 
         \hline
         1 & 92.2 & 60.0 & 63.6 & 73.8 & 107.0\\
         2 & 92.4 & 60.2 & 63.5 & 73.5 & 106.9\\
         3 & 92.7 & 60.7 & 63.4 & 73.4 & 106.8\\
         \hline
         4 & 92.9 & 60.7 & 63.1 & 73.2 & 106.6\\
         \hline
         5 & 92.7 & 60.4 & 63.5 & 73.5 & 106.9\\
         6 & 92.5 & 60.2 & 63.8 & 74.1 & 107.5\\
         7 & 92.1 & 60.0 & 63.9 & 74.5 & 107.9\\
            
         \bottomrule
    \end{tabular}%
    }
\label{tab:ab3-beta}
\end{table}

\begin{table}[!ht]
    \scriptsize
    \centering
    \caption{Parameter analysis of $\gamma$ on 3DHP and 3DPW}
    \resizebox{1.0\linewidth}{!}{%
    \begin{tabular}{c|c|c|c|c|c}
         \toprule
          & \multicolumn{3}{c|}{3DHP} & \multicolumn{2}{c}{3DPW} \\
         \hline 
         $\gamma$  & PCK $\uparrow$ & AUC $\uparrow$ & MPJPE $\downarrow$ & PA-MPJPE $\downarrow$ & MPJPE $\downarrow$  \\ 
         \hline
         0.7 & 92.4 & 60.0 & 63.8 & 74.1 & 107.5\\
         0.8 & 92.7 &  60.2 &  63.8 & 73.8 & 107.0\\
         0.9 & 92.9 &  60.5 &  63.3 & 73.5 & 106.9\\
         \hline
         1 & 92.9 &  60.7 &  63.1 & 73.2 & 106.6\\
         \hline
         1.1 & 92.7 &  60.7 &  63.6 & 73.8 & 107.0\\
         1.2 & 92.4 & 60.5 & 63.8 & 74.1 & 107.5\\
         1.3 & 92.2 & 60.0 & 63.9 & 74.4 & 107.7\\
            
         \bottomrule
    \end{tabular}%
    }
\label{tab:ab3-gamma}
\end{table}

\begin{table}[!ht]
    \scriptsize
    \centering
    \caption{Parameter analysis of $k$ on 3DHP and 3DPW}
    \resizebox{1.0\linewidth}{!}{%
    \begin{tabular}{c|c|c|c|c|c}
         \toprule
          & \multicolumn{3}{c|}{3DHP} & \multicolumn{2}{c}{3DPW} \\
         \hline 
         $k$  & PCK $\uparrow$ & AUC $\uparrow$ & MPJPE $\downarrow$ & PA-MPJPE $\downarrow$ & MPJPE $\downarrow$  \\ 
         \hline
         1 & 92.9 &  60.7 &  63.1 & 73.2 & 106.6\\
         \hline
         4 & 92.9 &  60.7 &  63.1 & 73.5 & 107.0\\
         7 & 92.5 & 60.4 & 63.4 & 73.8 & 107.3\\
         10 & 92.2 & 60.1 & 63.8 & 74.3 & 107.5\\
            
         \bottomrule
    \end{tabular}%
    }
\label{tab:ab3-k}
\end{table}

In Tabs. \ref{tab:ab3-alpha1} through \ref{tab:ab3-k}, we perform a parameter analysis on $\alpha_1$, $\alpha_2$, $\beta$, $\gamma$, and $k$ via the two cross-dataset evaluation tasks on 3DHP and 3DPW respectively. Notably, considering the analogous functions of $\beta_1$ and $\beta_2$ in serving as trade-off parameters in the differential discrimination, we consolidate them into a single parameter denoted as $\beta$ in Tab. \ref{tab:ab3-beta}. These results not only validate the appropriateness of our parameter choices but also demonstrate the stability of our proposed framework across varying parameter settings.

\section{Multi-frame Domain Generalization for 3D Human Pose Estimation}
\label{sec:multi}

While our proposed framework is specifically tailored for single-frame tasks in 3D Human Pose Estimation (HPE), such as PoseAug \cite{gong2021poseaug}, CEE-Net \cite{li2023cee}, and PoseDA \cite{chai2023poseda}, there are alternative approaches that extend their considerations to multi-frame settings through temporal-based techniques, exemplified by PoseAug-V \cite{zhang2023poseaug} and DH-AUG \cite{huang2022dh}. In this section, we present a comparative analysis of our method against these temporal-based multi-frame approaches within the 27-frame setting. Notably, our method remains consistent with the single-frame setting, except for variations in input size that contain 27 frames.

\begin{table}[!ht]
    \scriptsize
    \centering
    \caption{Cross-dataset evaluation with  MPJPE ($\downarrow$) and PA-MPJPE ($\downarrow$) on 3DHP and 3DPW (27-frame). }
    \resizebox{1.0\linewidth}{!}{%
    \begin{tabular}{c|c|c|c|c|c}
          \toprule
          \multicolumn{2}{c|}{} & \multicolumn{2}{c|}{3DHP} & \multicolumn{2}{c}{3DPW} \\
          \hline
          Method  & DG & MPJPE $\downarrow$ & PA-MPJPE $\downarrow$ & MPJPE $\downarrow$ & PA-MPJPE $\downarrow$  \\
         \hline
         VPose \cite{pavllo20193d} & $\times$ & 96.4 & 66.5 & 103.3 & 63.6\\
         VPose + PoseAug-V \cite{zhang2023poseaug}& $\checkmark$ & 86.5 & 61.0 & 91.1 & 54.3\\
         VPose + DH-AUG \cite{huang2022dh} & $\checkmark$ & 80.9 & 58.6 & 87.3 & 52.5\\
         VPose + Ours & $\checkmark$ & \textbf{79.7} & \textbf{57.9} & \textbf{85.2} & \textbf{51.6}\\
         \hline
         PoseFormer \cite{zheng20213d} & $\times$ & 93.3 & 66.7 & 118.5 & 73.4\\
         PoseFormer + PoseAug-V \cite{zhang2023poseaug}& $\checkmark$ & 82.9 & 63.1 & 108.3 & 64.8\\
         PoseFormer + DH-AUG \cite{huang2022dh} & $\checkmark$ & 75.4 & 61.8 & 104.4 & 62.1\\
         PoseFormer + Ours & $\checkmark$ & \textbf{74.1} & \textbf{61.0} & \textbf{102.7} & \textbf{61.3}\\


         \bottomrule
    \end{tabular}%
    }
\label{tab:multi}
\end{table}

In Tab. \ref{tab:multi}, our single-frame approach demonstrates a significant performance advantage over temporal-based methods like PoseAug-V and DH-AUG, despite being a single-frame method without relying on temporal-based techniques. This result underscores the effectiveness and superiority of our proposed method.

\section{Visualizations of Distributions between Source Data and Synthesized Data}

\label{sec:vis}
\begin{figure}[!ht]
  \centering
   \includegraphics[width=1.00\linewidth]{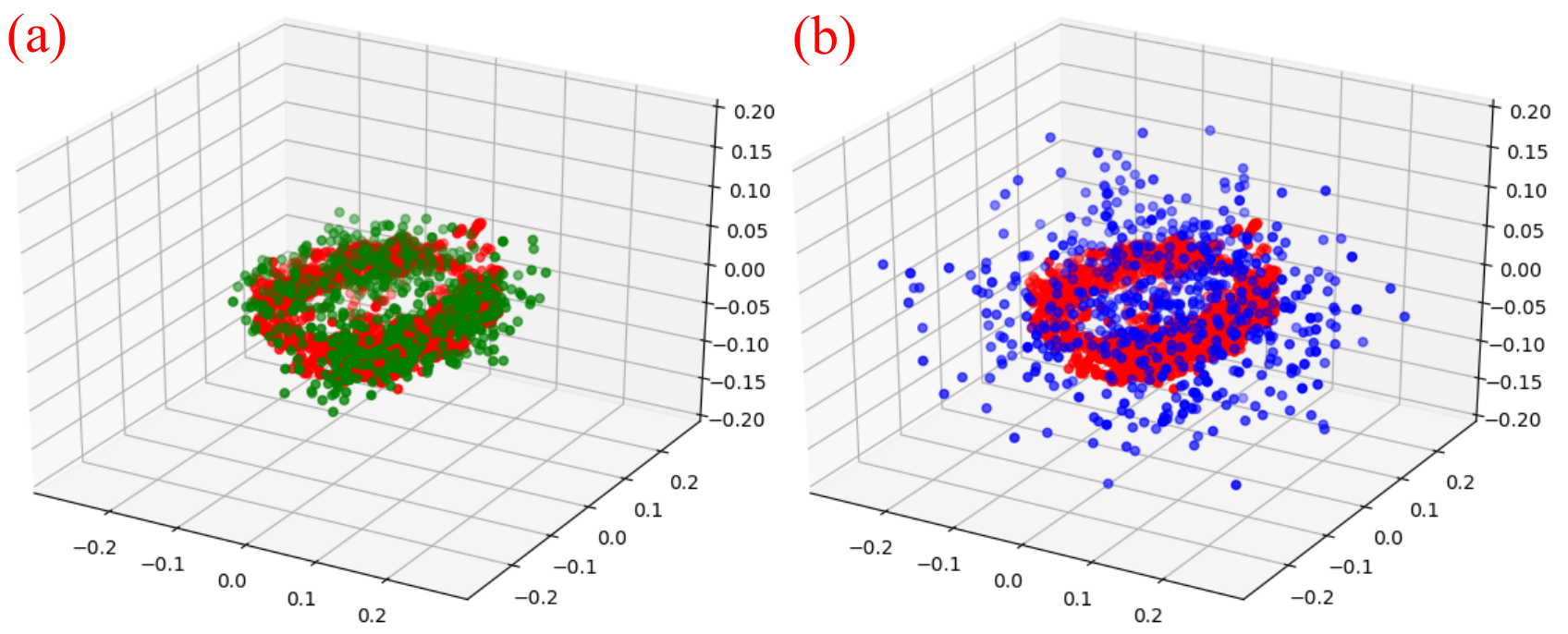}

   \caption{ Visualizations of Distributions. Here we choose the right hip joint in S6 set of Human3.6M for illustration. Red dots in both figures are the source data's distributions. Green dots on the left side are 3K points generated by the weak-augmentor, and blue dots on the right side are 3K points generated by the strong-augmentor.
   } %
   \label{fig:dist}
\end{figure}
\begin{figure}[!ht]
  \centering
   \includegraphics[width=1.00\linewidth]{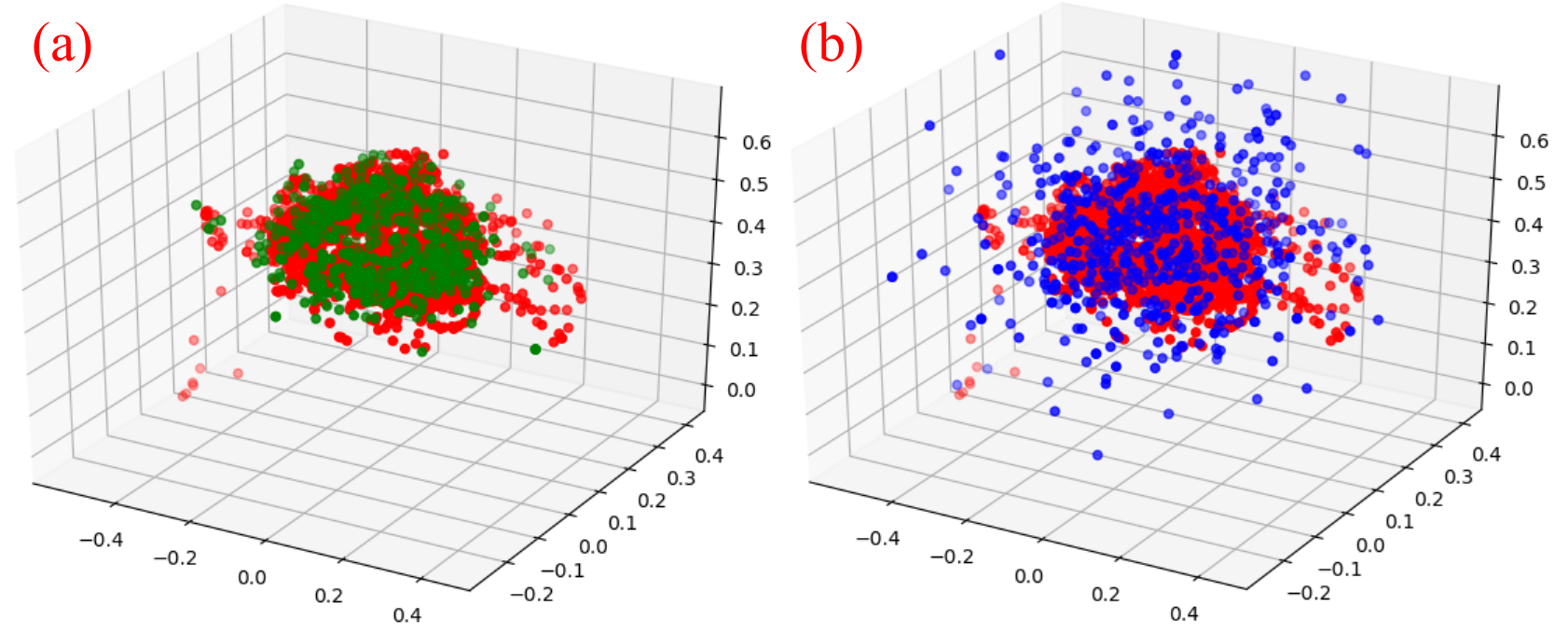}

   \caption{ Visualizations of Distributions. Here we choose the right shoulder joint in S6 set of Human3.6M for illustration. Red dots in both figures are the source data's distributions. Green dots on the left side are 3K points generated by the weak-augmentor, and blue dots on the right side are 3K points generated by the strong-augmentor.
   } %
   \label{fig:dist_2}
\end{figure}

In Fig. \ref{fig:dist} and Fig. \ref{fig:dist_2}, we present a qualitative visualization of our augmented data. We utilize S6 from Human3.6M \cite{ionescu2013human3} as the source domain (depicted by red dots in both figures). Subsequently, we generate 3,000 data points using both the weak augmentor (depicted by green dots on the left side) and the strong augmentor (depicted by blue dots on the right side). 

In these two figures, it is evident that data points generated by the weak augmentor closely resemble the distributions of the source, whereas those generated by the strong augmentor exhibit a notable deviation from the source distributions, thereby demonstrating the effectiveness of these two augmentors.

\section{Visualizations of Weak-augmented and Strong-augmented Poses}
\label{sec:cmp}

\begin{figure}[!ht]
  \centering
   \includegraphics[width=1.00\linewidth]{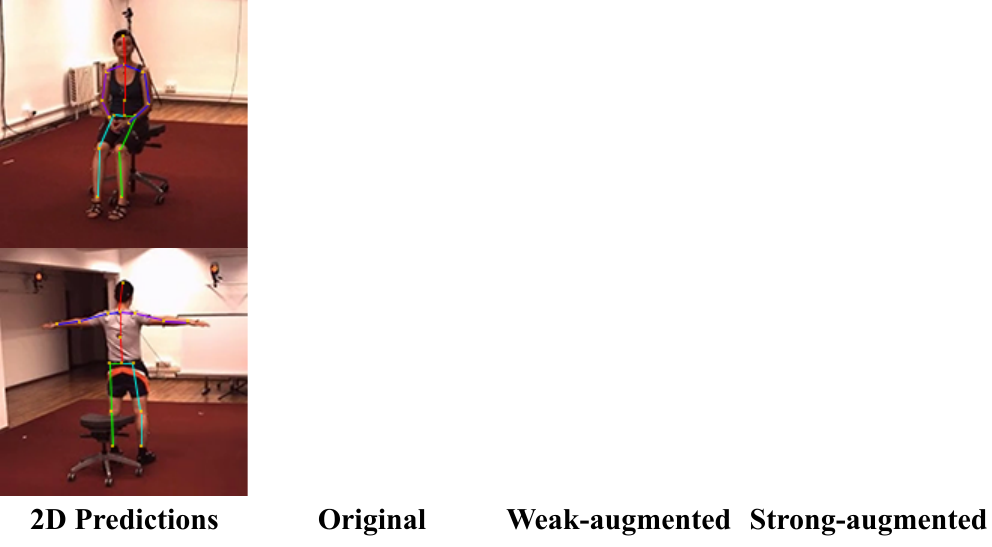}

   \caption{ Visualizations of Augmentations. 
   } %
   \label{fig:cmp}
\end{figure}

In Fig. \ref{fig:cmp}, we provide several examples to visualize the effectiveness of our dual-augmentor system. 

As depicted in this figure, the poses generated by the weak augmentor exhibit similarity to the original source poses. In contrast, poses generated by the strong augmentor differ significantly from the source poses. Nevertheless, these strong-augmented poses remain reasonable in their respective scenarios and align with the human model.

\section{Details of Laplacian Weighted Similarity}
\label{sec:laplacian}

\begin{figure}[!ht]
  \centering
   \includegraphics[width=0.5\linewidth]{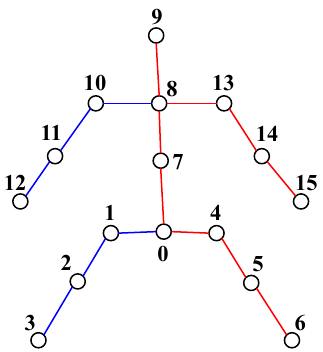}

   \caption{ Visualizations of 16-joint Human Body Model. 
   } %
   \label{fig:body}
\end{figure}

In Fig. \ref{fig:body}, the 16-joint human body model is provided, which follows the settings of previous works \cite{zhang2023poseaug, gholami2022adaptpose,li2023cee,huang2022dh,chai2023poseda,liu2023posynda}. Based on this model, it is straightforward to obtain the two matrices, the adjacency matrix $A$:
\begin{equation}
     A =
  \begin{bmatrix}
    0 & 1 & 0 & 0 & 1 & 0 & 0 & 1 &\hdots\\
    1 & 0 & 1 & 0 & 0 & 0 & 0 & 0 &\hdots\\
    0 & 1 & 0 & 1 & 0 & 0 & 0 & 0 &\hdots\\
    \vdots & \vdots & \vdots & \vdots & \vdots & \vdots & \vdots & \vdots & \ddots\\
  \end{bmatrix}_{16\times16},
\end{equation} and the degree matrix $D$:

\begin{equation}
     D =
  \begin{bmatrix}
    3 & & & &\\
    & 2 & & &\\
    & & 2 & &\\
    & & & \ddots &\\
    & & & & 1
  \end{bmatrix}_{16\times16}
\end{equation}


\end{document}